\documentclass[letterpaper, 10pt, conference]{ieeeconf}      

\IEEEoverridecommandlockouts                              

\usepackage{cite}
\usepackage{smartdiagram}
\usepackage{amsmath,amssymb,amsfonts}
\usepackage{algorithmic}
\usepackage{graphicx}
\usepackage{textcomp}
\usepackage{xcolor}
\usepackage{tikz}
\usepackage{capt-of} 
\usepackage[font=small]{caption}
\usepackage{csquotes}
\usepackage{etoolbox}
\usepackage{siunitx}
\usepackage{hyperref}
\usepackage{url}
\usetikzlibrary{shapes.geometric, arrows}
\tikzstyle{startstop} = [rectangle, rounded corners, minimum width=3cm, minimum height=1cm, text centered, draw=black, fill=teal!60]
\tikzstyle{arrow} = [thick,->,>=stealth]

\usepackage{subfigure}

\DeclareMathOperator*{\argmin}{arg\!\min}

\renewcommand{\baselinestretch}{.89} 

\usepackage{acro}
\newcommand{\newac}[2]{\DeclareAcronym{#1}{short=#1,long=#2}}
\newac{ADL}{Activities of Daily Living}
\newac{BCI}{Brain Computer Interface}
\newac{BMI}{Brain Machine Interface}
\newac{CoM}{Center of Mass}
\newac{CS}{Constant Strain}
\newac{COM}{Center of Mass}
\newac{DOF}{Degrees of Freedom}
\newac{EEG}{Electroencephalography}
\newac{EOM}{Equations of Motion}
\newac{ERD}{Event-Related Desynchronization}
\newac{ERS}{Event-Related Synchronization}
\newac{GBN}{Generalized Binary Noise}
\newac{HMI}{Human Machine Interface}
\newac{HSA}{Handed Shearing Auxetic}
\newac{LDA}{Linear Discriminant Analysis}
\newac{PCS}{Piecewise Constant Strain}
\newac{RMSE}{Root Mean-Squared Error}

\definecolor{adaee4ed-88c8-5b21-a9e7-31316ebef86f}{RGB}{255, 179, 178}
\definecolor{f3551e38-74df-57e2-b793-83d7fe876c85}{RGB}{0, 0, 0}
\definecolor{0b71a967-1f15-55a5-9bb9-70efa7b4fc58}{RGB}{51, 51, 51}
\definecolor{58712c6c-1b6d-5716-9834-102aad6341dc}{RGB}{175, 179, 255}
\definecolor{70af333d-4869-5f2a-97ca-9904a9fce6c3}{RGB}{179, 254, 174}
\definecolor{747aec21-333b-59ee-84e3-ddff893e5ccd}{RGB}{255, 216, 176}
\definecolor{5856d031-3da1-575c-834e-c77e9e438c62}{RGB}{162, 177, 195}

\definecolor{e4a2dd21-3fe7-5275-8b0b-724ada9b9d47}{RGB}{195, 179, 255}
\definecolor{f3551e38-74df-57e2-b793-83d7fe876c85}{RGB}{0, 0, 0}
\definecolor{0b71a967-1f15-55a5-9bb9-70efa7b4fc58}{RGB}{51, 51, 51}
\definecolor{78fe7373-ff9c-5966-abea-d218736ba889}{RGB}{101, 215, 230}
\definecolor{3a27046a-3c72-57cf-88d2-39ae37b9c0a6}{RGB}{221, 255, 179}
\definecolor{a4696e52-3ed0-53e9-9d72-99793a92efef}{RGB}{244, 168, 62}
\definecolor{5856d031-3da1-575c-834e-c77e9e438c62}{RGB}{162, 177, 195}

\tikzstyle{start} = [rectangle, rounded corners, minimum width=3cm, minimum height=1cm, text centered, font=\normalsize, color=0b71a967-1f15-55a5-9bb9-70efa7b4fc58, draw=f3551e38-74df-57e2-b793-83d7fe876c85, line width=1, fill=adaee4ed-88c8-5b21-a9e7-31316ebef86f]
\tikzstyle{training} = [rectangle, rounded corners, minimum width=3cm, minimum height=1cm, text centered, font=\normalsize, color=0b71a967-1f15-55a5-9bb9-70efa7b4fc58, draw=f3551e38-74df-57e2-b793-83d7fe876c85, line width=1, fill=58712c6c-1b6d-5716-9834-102aad6341dc]
\tikzstyle{testing} = [rectangle, rounded corners, minimum width=3cm, minimum height=1cm, text centered, font=\normalsize, color=0b71a967-1f15-55a5-9bb9-70efa7b4fc58, draw=f3551e38-74df-57e2-b793-83d7fe876c85, line width=1, fill=58712c6c-1b6d-5716-9834-102aad6341dc]
\tikzstyle{decision} = [diamond, minimum width=3cm, minimum height=2cm, text centered, font=\normalsize, color=0b71a967-1f15-55a5-9bb9-70efa7b4fc58, draw=f3551e38-74df-57e2-b793-83d7fe876c85, line width=1, fill=70af333d-4869-5f2a-97ca-9904a9fce6c3]
\tikzstyle{action} = [rectangle, rounded corners, minimum width=3cm, minimum height=1cm, text centered, font=\normalsize, color=0b71a967-1f15-55a5-9bb9-70efa7b4fc58, draw=f3551e38-74df-57e2-b793-83d7fe876c85, line width=1, fill=747aec21-333b-59ee-84e3-ddff893e5ccd]
\tikzstyle{arrow} = [thick, draw=5856d031-3da1-575c-834e-c77e9e438c62, line width=2, ->, >=stealth]
\tikzstyle{10575355-9019-58aa-a2fd-04c41c41d200} = [rectangle, minimum width=3cm, minimum height=1cm, text centered, font=\normalsize, color=0b71a967-1f15-55a5-9bb9-70efa7b4fc58, draw=f3551e38-74df-57e2-b793-83d7fe876c85, line width=1, fill=e4a2dd21-3fe7-5275-8b0b-724ada9b9d47]
\tikzstyle{5b9e070c-fb6f-5d55-96a4-e63717baaf1e} = [rectangle, minimum width=3cm, minimum height=1cm, text centered, font=\normalsize, color=0b71a967-1f15-55a5-9bb9-70efa7b4fc58, draw=f3551e38-74df-57e2-b793-83d7fe876c85, line width=1, fill=78fe7373-ff9c-5966-abea-d218736ba889]
\tikzstyle{1e99a6d6-ba21-5a81-a188-71beac317701} = [rectangle, minimum width=3cm, minimum height=1cm, text centered, font=\normalsize, color=0b71a967-1f15-55a5-9bb9-70efa7b4fc58, draw=f3551e38-74df-57e2-b793-83d7fe876c85, line width=1, fill=3a27046a-3c72-57cf-88d2-39ae37b9c0a6]
\tikzstyle{0c48bdee-3f98-522c-85e7-bc0cc3372c62} = [rectangle, minimum width=3cm, minimum height=1cm, text centered, font=\normalsize, color=0b71a967-1f15-55a5-9bb9-70efa7b4fc58, draw=f3551e38-74df-57e2-b793-83d7fe876c85, line width=1, fill=a4696e52-3ed0-53e9-9d72-99793a92efef]
\tikzstyle{7be24b85-97d0-5b76-ba9e-d94005dca8f2} = [thick, draw=5856d031-3da1-575c-834e-c77e9e438c62, line width=2, ->, >=stealth]

\def\BibTeX{{\rm B\kern-.05em{\sc i\kern-.025em b}\kern-.08em
    T\kern-.1667em\lower.7ex\hbox{E}\kern-.125emX}}

\title{\Large \bf
Guiding Soft Robots with Motor-Imagery Brain Signals and Impedance Control
}

\author{Maximilian Stölzle$^{*, 1}$, Sonal Santosh Baberwal$^{*,2}$, Daniela Rus$^{3}$, Shirley Coyle$^{\dagger,2}$, and Cosimo Della Santina$^{\dagger,1}$ 
\thanks{$^*$Authors contributed equally, $^\dagger$Authors supervised equally.}
\thanks{The work by Maximilian Stölzle was supported under the European Union's Horizon Europe Program from Project EMERGE - Grant Agreement No. 101070918. The work by SS. Baberwal was supported by a grant from Science Foundation Ireland under Grant numbers 18/CRT/6183, SFI/12/RC/2289 P2}
\thanks{$^{1}$M. Stölzle, and C. Della Santina are with the Cognitive Robotics department, Delft University of Technology, Mekelweg 2, 2628 CD Delft, Netherlands {\tt\scriptsize \{M.W.Stolzle, C.DellaSantina\}@tudelft.nl}.}
\thanks{
$^{2}$SS. Baberwal and S. Coyle are with the School of Electronics Engineering, Dublin City University, Dublin, Ireland, {\tt\scriptsize sonal.baberwal2@mail.dcu.ie, shirley.coyle@dcu.ie}.}
\thanks{
$^{3}$D. Rus is with the MIT Computer Science and Artificial Intelligence Laboratory (CSAIL), Massachusetts Institute of Technology, Cambridge, MA 02139 USA, {\tt\scriptsize rus@csail.mit.edu}.
}
}

\begin{document}

\bstctlcite{IEEEexample:BSTcontrol}

\maketitle
\thispagestyle{empty}
\pagestyle{empty}

\begin{abstract}
%
Integrating Brain-Machine Interfaces into non-clinical applications like robot motion control remains difficult - despite remarkable advancements in clinical settings. Specifically, EEG-based motor imagery systems are still error-prone, posing safety risks when rigid robots operate near humans. This work presents an alternative pathway towards safe and effective operation by combining wearable EEG with physically embodied safety in soft robots. We introduce and test a pipeline that allows a user to move a soft robot's end effector in real time via brain waves that are measured by as few as three EEG channels. A robust motor imagery algorithm interprets the user's intentions to move the position of a virtual attractor to which the end effector is attracted, thanks to a new Cartesian impedance controller. We specifically focus here on planar soft robot-based architected metamaterials, which require the development of a novel control architecture to deal with the peculiar nonlinearities - e.g., non-affinity in control. We preliminarily but quantitatively evaluate the approach on the task of setpoint regulation. We observe that the user reaches the proximity of the setpoint in 66\% of steps and that for successful steps, the average response time is 21.5s. We also demonstrate the execution of simple real-world tasks involving interaction with the environment, which would be extremely hard to perform if it were not for the robot's softness.


\end{abstract}


\section{Introduction}
\acp{BMI} \cite{liu2024cognitive} facilitate the translation of neural activity into actionable commands, enabling individuals to control external devices and systems through their thoughts and attention~\cite{coyle2007brain,lee2017brain}. Compared to traditional bulky EEG setups~\cite{van2012brain}, one of the emerging avenues towards practical and wearable \acp{EEG} devices are systems based on motor imagery signals due to their intuitiveness and no external dependency on (e.g., visual) stimuli. These have been used in stroke rehabilitation \cite{khan2020review}. Several works in literature have considered using this technology to control robot manipulators~\cite{schiatti2017soft,aldini2019effect,zhang2023noir}. 

However, the state-of-the-art classifiers on few-channel, online \ac{EEG} signals are still limited in achieving an accuracy of 65-75 \si{\percent}~\cite{arpaia2022non, zhang2023noir} and are prone to producing outliers, which make it very challenging to operate robots safely and robustly using these techniques~\cite{liu2024cognitive}. In (rigid) robotics literature, this has been addressed by relying on force-based (i.e., impedance) control~\cite{schiatti2017soft} and by making the robot's behavior more predictable~\cite{aldini2019effect}.
%
%
In this work, we follow a different path and investigate \textit{embodying} safety by pairing soft robots~\cite{rus2015design, della2020soft} 
with \ac{BMI}. This way, risks can be mitigated, and more natural interactions with an unstructured environment can be achieved by relying on structural compliance.

\begin{figure*}
    \centering
    \includegraphics[width=0.95\textwidth]{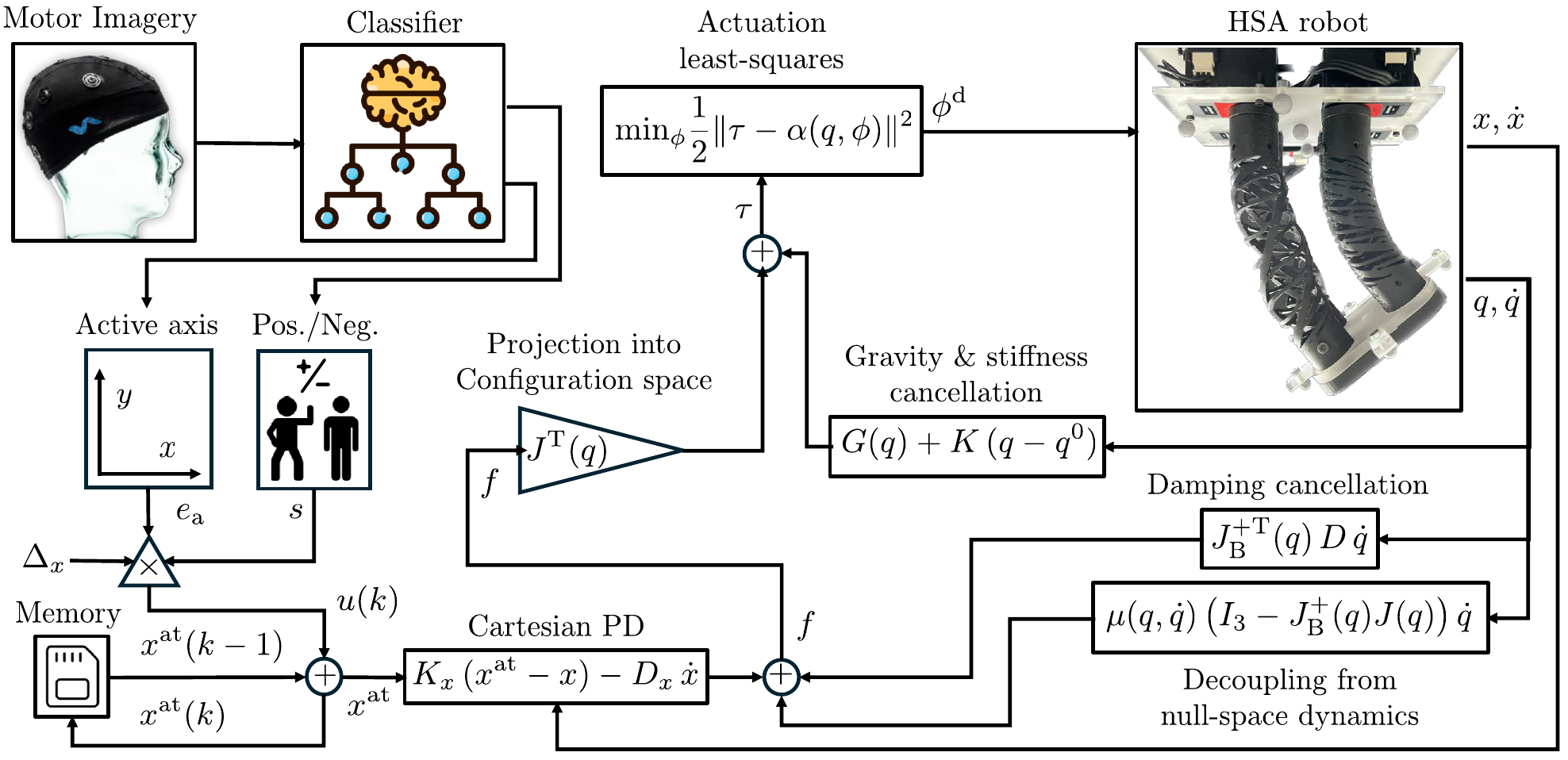}
    \caption{Scheme of the proposed approach to control HSA robots with motor imagery. Brain signals steer an attractor in operational space: first, we switch the active coordinate axis when we detect jaw clenching. If no jaw clenching is detected, we classify the EEG signals based on left/right motor imaginations into positive and negative movements along the active axis. Next, we regulate the robot towards the chosen attractor position $x^\mathrm{at}$ with a Cartesian impedance controller. This controller first cancels all static forces and the residual coupling of the null space on the operational space dynamics. This allows us to now shape our own potential with a PD term in operational space. As the robot is underactuated, we optimize least-squares to identify the actuation $\phi^\mathrm{d}$ so that the residual between the desired and actual torques in the configuration space is minimized. Icons created by Flaticon\copyright.}
    \label{fig:control_scheme}
\end{figure*}

While \ac{BMI}-based assistance has been investigated with a focus on 
soft exosuits assisting hand-rehabilitation ~\cite{zhang2019eeg} or with strenuous arm acitivites~\cite{tacca2022neuro}, 
it is still an open challenge how \ac{BMI} can be used for controlling soft manipulators. 
In this work, we make a first step towards solving this challenge by proposing a pipeline (see Fig. \ref{fig:control_scheme}) that lets the user steer the soft robot's end-effector in Cartesian space. The two key ingredients are a novel mapping strategy transforming the brain signals into meaningful references and Cartesian impedance control. The latter is essential because it allows for preserving the robot's compliance in closed-loop \cite{della2017controlling}. We build the proposed \ac{BMI} pipeline around a \ac{HSA} soft robot \cite{stolzle2023modelling, stolzle2023experimental}, 
which relies on architected metamaterials and electrical actuation to elongate, bend, and twist. This makes the control problem especially challenging because of the peculiarity of these systems' dynamics, namely, underactuation and non-affinity in control. We provide more details on innovation from the model-based control standpoint in Section~\ref{sub:computational_controller}.
 
We quantitatively verify the entire approach on mind-controlled setpoint regulation involving tracking a reference consisting of nine-step functions and demonstrate the qualitative behavior when assisting with a simple daily living activity. 
Furthermore, we compare the performance of our motor imagery-based control to approaches giving the computational controller access to the privileged information of setpoints, which can be considered to be an upper bound on performance.

Our contributions are: (i) Establishing a \ac{BMI} strategy for continuum soft robots. This strategy is supported by experiments in which we perform setpoint regulation with a planar \ac{HSA} robot and motor imagery, (ii) A Cartesian impedance controller for \ac{HSA} robots, which we experimentally validated on a simple \ac{ADL} task involving environment interaction: the user needs to steer the end-effector towards the tip of a hairspray container, apply force for releasing the fluid, and finally let the robot retreat from the contact.          

A video attachment to this paper, including recordings of experimental results, can be found on YouTube\footnote{\url{https://youtu.be/wZTOxBPZmPc}}.
Furthermore, we have open-sourced our code, including the OpenVibe pipeline, on GitHub\footnote{\url{https://github.com/tud-phi/sr-brain-control}}.

\section{Task-space Impedance Control}
In this work, we let the user steer with motor imaginary brain signals the Cartesian position $x \in \mathbb{R}^2$ of the end-effector (i.e., the platform) of a planar \ac{HSA} robot.
We realize this strategy by first classifying the motor imaginary signals into Cartesian-space movement directions (e.g., the active axis and sign of the movement). We use this information to adjust the position of a task-space attractor iteratively (see Section~\ref{sub:planning_attractors_switching}). Section~\ref{sub:computational_controller} describes how a model-based computational controller establishes this attractor. Importantly, we preserve the soft robot's compliance by shaping the closed-loop system's impedance in Cartesian space.

\subsection{Background: Motor Imagery-based BMI systems}\label{sub:motor_imagery_bmi}

Imagining the movement of body parts or limbs (e.g., hands, legs, tongue) without moving it or the mental rehearsal of a motor act without overt movement execution is termed Motor Imagery~\cite{lotze2006motor}.  The neuronal activities observable inside a frequency range of \SI{8}{Hz} to \SI{12}{Hz} (Mu) and \SI{12}{Hz} to \SI{30}{Hz} (Beta) are associated with cortical areas directly connected to the brain’s motor output (activating primary sensorimotor areas that can be modulated with imaginary mental movement in healthy as well people with neuromuscular disabilities). 

The motor imagery \ac{BMI} framework typically consists of four integral components:

\begin{enumerate}
    \item \textbf{Signal acquisition:} The initial stage involves the recording of neural signals while the person imagines the movements of the limbs, generally acquired using noninvasive methodologies (e.g., \ac{EEG}).
    \item \textbf{Feature extraction:} Following signal acquisition, signal processing techniques are applied to extract salient features from the neural patterns associated with specific cognitive processes or intentions.
    \item \textbf{Feature translation:} This translation phase interprets the user's cognitive intent, converting it into actionable instructions for external devices.
    \item \textbf{Device output:} The culmination of the \ac{BMI} process is the application of the interpreted commands to external devices. 
\end{enumerate}

As detailed further in Sec.~\ref{sub:bmi_protocol}, we leverage the difference in signals when imagining motor actions vs. the rest state to control the sign of movement. The active axis of movement can be switched by clenching the jaw. 


\subsection{Planning attractors with brain signals}\label{sub:planning_attractors_switching}
Our brain signal processing pipeline provides us with two pieces of information at each time step $k$: i) the unit vector $e_\mathrm{a}(k) \in \{ [1, 0]^\mathrm{T}, [0, 1]^\mathrm{T} \}$ corresponding to the current active axis of movement 
and ii) the sign of movement $s(k) \in \{ -1, 1 \}$. We use $e_\mathrm{a}(k)$ and $s(k)$ to incrementally steer a virtual attractor defined in operational space $x^\mathrm{at} \in \mathbb{R}^2$ as follows
%
\begin{equation}\small
    u(k) = s(k) \, e_\mathrm{a}(k) \in \mathbb{R}^2, \quad  x^\mathrm{at}(k) = x^\mathrm{at}(k-1) + \Delta_\mathrm{x} \, u(k),
\end{equation}
where $\Delta_\mathrm{x} \in \mathbb{R}^+$ is a tunable constant influencing the velocity of the attractor movement.
Later, we will shape the potential field with a computational controller such that the attractor becomes a globally asymptotically stable equilibrium (see Section~\ref{sub:computational_controller}).

\begin{figure}
\begin{center}
    \includegraphics[width=\columnwidth]{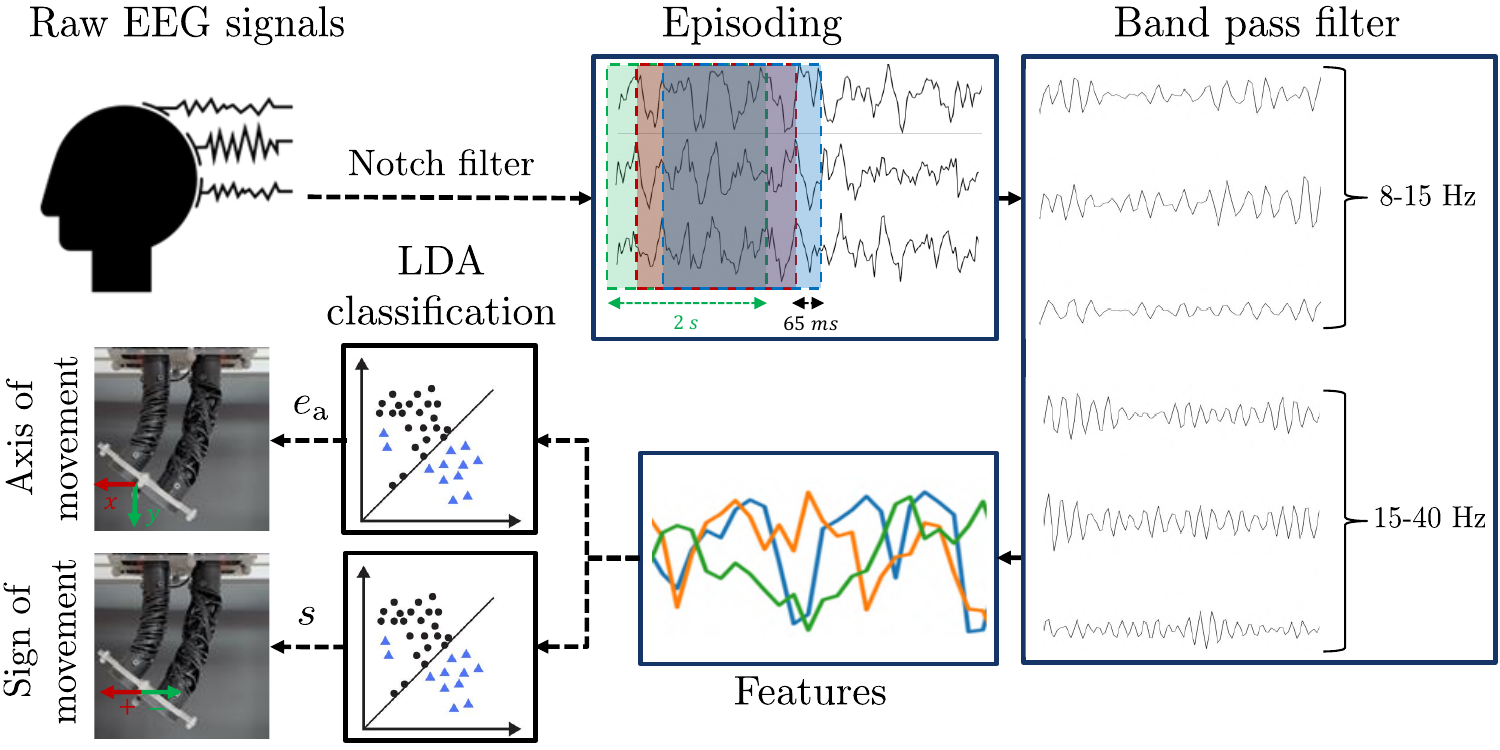}
    \caption{EEG data processing pipeline: The EEG data is acquired in real-time, pre-processed, and divided into episodes and subbands. Next, we extract power features and pass them to two LDA classifiers: the first outputs the axis of movement (for example, moving along the x- or y-axis), and the second provides the sign of movement (for example, positive or negative movement along the active axis). These commands are then used to move the attractor in Cartesian space.}
    \label{fig:eeg_pipeline}
\end{center}
\end{figure}

\subsection{Background: modeling planar HSA robots}
Robots based on \acp{HSA} rely on rods made of architected metamaterials to generate motion.  
More specifically, twisting the rods along their handedness leads to an elongation of the rod~\cite{good2022expanding}. Combining multiple \acp{HSA} in the setting of a parallel robot and actuating them with servo motors allows us to generate complex motion primitives and offer beneficial mechanical characteristics such as a high stiffness-to-weight ratio~\cite{stolzle2023modelling, good2022expanding}.

Prior work~\cite{stolzle2023experimental} has shown that the shape of planar \ac{HSA} robots can be approximated by one \ac{CS} segment. Therefore, we define the configuration of the system as $q = \begin{bmatrix}
    \kappa_\mathrm{be} & \sigma_\mathrm{sh} & \sigma_\mathrm{ax}
\end{bmatrix}^\mathrm{T} \in \mathbb{R}^3$.
We also have access to closed-form formulations of the forward kinematics $\pi: q \rightarrow \chi$ and inverse kinematics $\varrho: \chi \rightarrow q$ where $\chi = \begin{bmatrix}
    x^\mathrm{T} & \theta
\end{bmatrix}^\mathrm{T} \in SE(2)$ is the pose in task space and $\theta$ represents the end-effector orientation~\cite{stolzle2023experimental}.
We use the notation $J(q) = \frac{\partial x}{\partial q} \in \mathbb{R}^{2\times3}$ to refer to the kinematic Jacobian.

We can then derive the dynamics of the planar \ac{HSA} robot in Euler-Lagrangian form
\begin{equation}\small\label{eq:configuration_space_dynamics}
    M(q) \Ddot{q} + C(q,\dot{q})\dot{q} + G(q) + K \, (q-q^0) + D \, \dot{q} = \alpha(q,\phi),
\end{equation}
where $M(q), C(q,\dot{q}) \in \mathbb{R}^{3 \times 3}$ captures the inertial and Coriolis effects, $G(q) \in \mathbb{R}^3$ contributes the gravitational forces and $K \in \mathbb{R}^{3 \times 3}$ is the stiffness of the robot in its un-actuated state $q^0$. Furthermore, $D \in \mathbb{R}^{3 \times 3}$ is a positive-definite damping matrix. Finally, for the planar case, two \ac{HSA} rods are assumed to be actuated by the motor/twist angle $\phi \in \mathbb{R}^2$. As the handedness of the rods will be accounted for later, we state the actuation bounds as $0 \leq \phi_i \leq \phi_\mathrm{max} \: \forall i \in \{ 1, 2 \}$.
The auxetic trajectory~\cite{good2022expanding} causes the motors to act through the elasticity of the rods on the system and modify the axial rest length of the rod as a function of the twist strain~\cite{stolzle2023modelling}.
Furthermore, the stiffness of the rod can be modeled to be an affine function with respect to the twist strain~\cite{good2022expanding, stolzle2023modelling}. Both effects are captured in the actuation function $\alpha(q,\phi)$, which is nonlinear with respect to the actuation coordinate $\phi$ and affine in the configuration $q$.
Although this has never been done in the context of HSA robots, it is immediate to see that their dynamics \eqref{eq:configuration_space_dynamics} can be projected into operational space, yielding the form~\cite{della2019exact, della2020model} 
\begin{equation}\small\label{eq:operational_space_dynamics}
    \Lambda(q) \, \Ddot{x} + \mu(q,\dot{q}) \dot{q} + J_\mathrm{B}^{+\mathrm{T}} ( G(q) + K (q-q^0) + D \, \dot{q} ) = J_\mathrm{B}^{+\mathrm{T}} \alpha(q,\phi),
\end{equation}
where $J_\mathrm{B}^+(q) = B^{-1}J^\mathrm{T}(J B^{-1} J^\mathrm{T})^{-1} \in \mathbb{R}^{3\times2}$ is the dynamically consistent pseudo-inverse, $\Lambda(q) = (J \, B^{-1} J^\mathrm{T})^{-1} \in \mathbb{R}^{2 \times 2}$ is the inertia matrix in task space, and $\mu(q, \dot{q}) = \Lambda(q) \, (J B^{-1} C - \dot{J}) \in \mathbb{R}^{2 \times 3}$ collects the Cartesian Coriolis and centrifugal terms. 

\subsection{Cartesian impedance controller}\label{sub:computational_controller}
In previous work~\cite{stolzle2023experimental}, we have devised a model-based control strategy for regulating a planar \ac{HSA} robot towards a desired position in task space. 
We introduce below a novel control strategy that addresses some limitations of our previous work that are critical for the \ac{BMI} application. Namely, we (i) avoid computationally demanding planning procedures, (ii) remove integral terms that are unsafe for environment interaction, and (iii) enable impedance shaping in operational space. This Cartesian-space impedance controller is inspired by ~\cite{ott2008cartesian,della2020model}, but specifically designed for and tailored to \ac{HSA} robots. Crucially, we need to overcome the challenges of underactuation and the nonlinearity in the actuation, which were not present in that original work. 

\subsubsection{Proposed controller}
We propose the following dynamic feedback law that renders $x^\mathrm{at}$ an attractor of the closed-loop system 
\begin{equation}\small\label{eq:cartesian_impedance_controller}
\begin{split}
    \tau =& J^\mathrm{T}(q) \, \left (K_x \, (x^\mathrm{at} - x) - D_x \, \dot{x} \right ) + G(q) + K \, (q-q^0)\\
    &+ J^\mathrm{T}(q) \, J_\mathrm{B}^{+\mathrm{T}}(q) \, D \, \dot{q} + J^\mathrm{T}(q) \, \mu(q,\dot{q}) \left ( I_3 - J_\mathrm{B}^+(q) J(q) \right )\dot{q}
\end{split}
\end{equation}
where $\tau \in \mathbb{R}^3$ is the desired torque in configuration space, $G(q) + K \, (q-q^0)$ cancels the acting gravitational and elastic forces, and $J^\mathrm{T} J_\mathrm{B}^{+\mathrm{T}} D \, \dot{q}$ removes the natural dissipation in operational space.
We emphasize that because the system is underactuated, we need to cancel the stiffness directly in the configuration instead of operational space as done in previous work~\cite{della2020model}.
We can shape our desired impedance characteristics in Cartesian space with the PD term $f_\mathrm{PD} = K_x \, (x^\mathrm{at} - x) - D_x \, \dot{x} $ which is then projected into configuration space by premultiplying with $J^\mathrm{T}(q)$.

The term $\mu(q,\dot{q}) \left ( I_3 - J_\mathrm{B}^+(q) J(q) \right )\dot{q}$ decouples the operational space dynamics from the residual of the null-space dynamics~\cite{della2020model}\cite[Ch. 4]{ott2008cartesian}.
The identity $\dot{q} = J_\mathrm{B}^+ \, \dot{x} + Z^\mathrm{T} \, \nu_\mathrm{N}$, where $Z^\mathrm{T} \in \mathbb{R}^{3 \times 1}$ is the dynamically-consistent pseudo-inverse of the null space, allows us to formulate $\dot{q}$ as a sum of the task-space velocity $\dot{x}$ and the null-space velocity $\nu_\mathrm{N}$. Leveraging this identity, the Coriolis and centrifugal matrix $\mu(q,\dot{q})$ can be split into a term $\mu_x(q,\dot{q}) = \mu \, J_\mathrm{B}^+ \in \mathbb{R}^{2 \times 2}$ excited by $x$ and the expression $\mu_\mathrm{N}(q,\dot{q}) = \mu \, Z^\mathrm{T} \in \mathbb{R}^{2 \times 1}$ that is excited by the null-space coordinates resulting in $\mu(q,\dot{q}) \, \dot{q} = \mu_x(q,\dot{q}) \, \dot{x} + \mu_\mathrm{N}(q,\dot{q}) \, \nu_\mathrm{N}$.
This allows us to cancel the term $\mu_\mathrm{N}(q,\dot{q}) \, \nu_\mathrm{N}$ through $\mu(q,\dot{q}) \left ( I_3 - J_\mathrm{B}^+(q) J(q) \right )\dot{q}$ without having to compute the null space explicitly.


In summary, the closed-loop dynamics in operational space can be stated as
\begin{equation}\small\label{eq:closed_loop_dynamics}
    \Lambda(q) \, \Ddot{x} + \mu(q,\dot{q}) \, J_\mathrm{B}^+ \, \dot{x} + K_x \, (x - x^\mathrm{at}) + D_x \, \dot{x} = 0,
\end{equation}
which results in $x^\mathrm{at}$ being the globally asymptotically stable equilibrium of the closed-loop operational space dynamics.
\footnote{Please note that this only holds under the assumption that the desired operational-space control input can be perfectly tracked by the actuators, which is generally not the case in underactuated settings.}


\begin{figure*}
\begin{center}
    \subfigure[Operational workspace]{\includegraphics[width=0.36\linewidth]{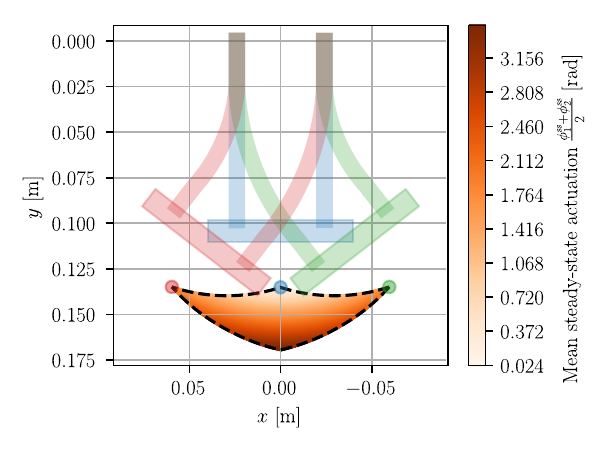}\label{fig:hsa_workspace}}
    \subfigure[Experimental setup]{\includegraphics[width=0.63\linewidth]{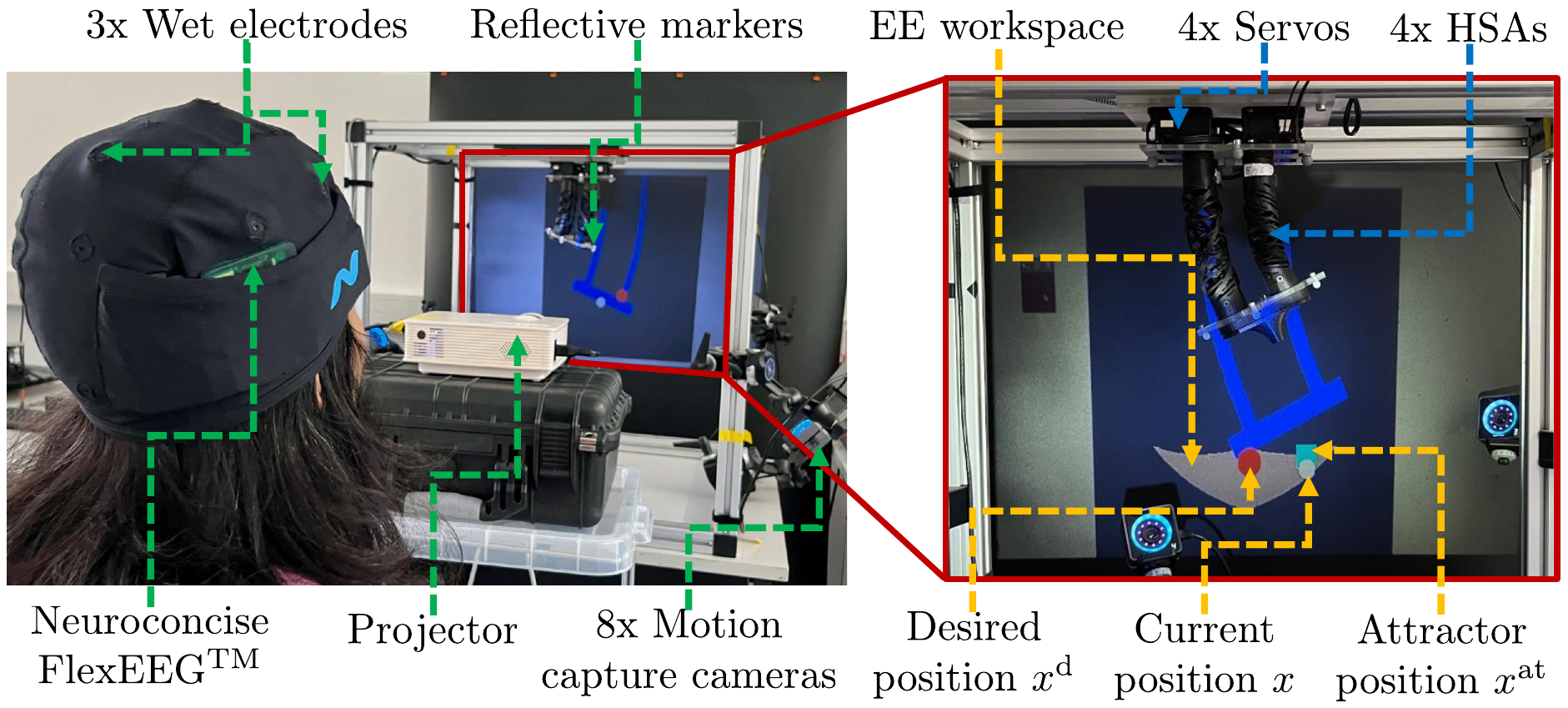}\label{fig:experimental_setup}}
    \caption{\textbf{Panel (a:} Operational workspace of an HSA robot with attached end-effector: the color displays the mean steady-state actuation $\frac{\phi_1^\mathrm{ss} + \phi_2^\mathrm{ss}}{2}$ necessary for the end-effector to remain at the position. Additionally, we visualize three example shapes: the straight configuration with $\phi^\mathrm{ss} = (0, 0)$ (blue), maximum clockwise bending with $\phi^\mathrm{ss} = (3.49, 0) \, \si{rad}$ (red), and maximum counter-clockwise bending with $\phi^\mathrm{ss} = (0, 3.49) \, \si{rad}$ (green).
    \textbf{Panel (b):} The HSA robot is mounted platform-down to a motion capture cage with 8x Optitrack PrimeX 13 cameras, which track the 3D pose of the platform (i.e., the end-effector). A Dynamixel MX-28 servo actuates each of the four HSAs. We project a rendering of the current (white dot) and desired (red dot) end-effector position, the attractor (green square), and the operational workspace (grey area) onto the black screen in the background. The study subject wears a cap with the Neuroconcise FlexEEG sensor, and we acquire the data of three electrodes connected to the motor cortex.}
\end{center}
\end{figure*}

\begin{figure}
\begin{center}
    \includegraphics[width=\columnwidth]{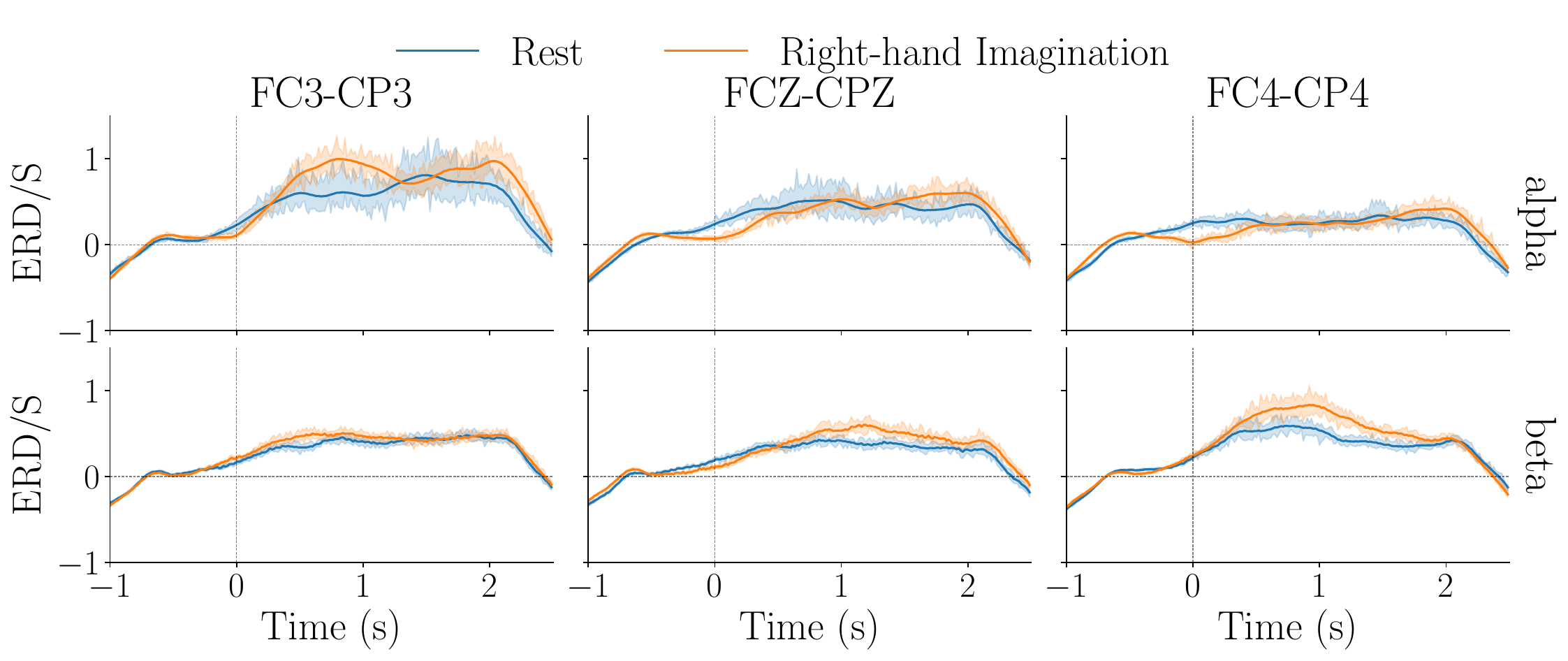}
    \caption{ERD/S (overall average) over a time period of \SI{2.5}{s} of training data for right-hand Imagination v/s rest state, including the Alpha and Beta bands of the EEG signals, where the cue is presented at \SI{0}{s}. We plot the data of three sensors (i.e., channels): FC3-CP3 (left), FCZ-CPZ (middle), and FC4-CP4 (right).}
    \label{fig:ERDS}
\end{center}
\end{figure}

\begin{figure*}[t]
    \centering
    \subfigure[End-effector x-coordinate]{\includegraphics[width=0.47\textwidth, trim={5, 5, 5, 5}]{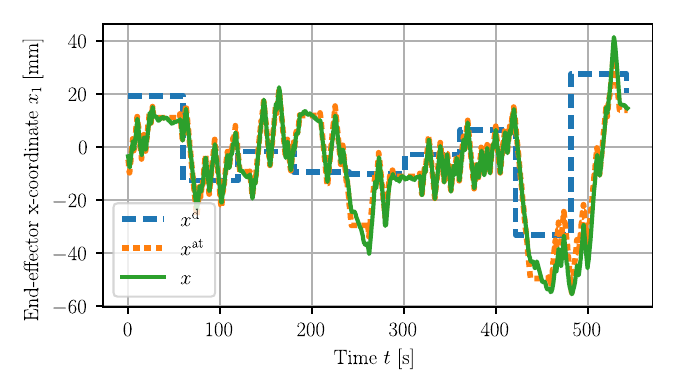}\label{fig:experimental_results:setpoint_regulation:brain:pee_x}}
    \subfigure[End-effector y-coordinate]{\includegraphics[width=0.47\textwidth, trim={5, 5, 5, 5}]{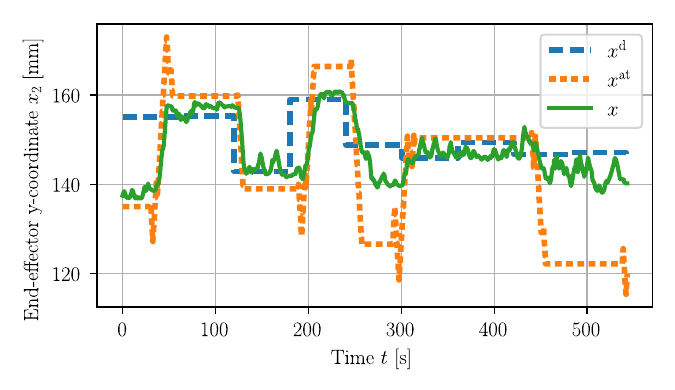}\label{fig:experimental_results:setpoint_regulation:brain:pee_y}}\\
    \vspace{-0.2cm}
    \subfigure[Configuration $q$]{\includegraphics[width=0.47\textwidth, trim={5, 5, 5, 5}]{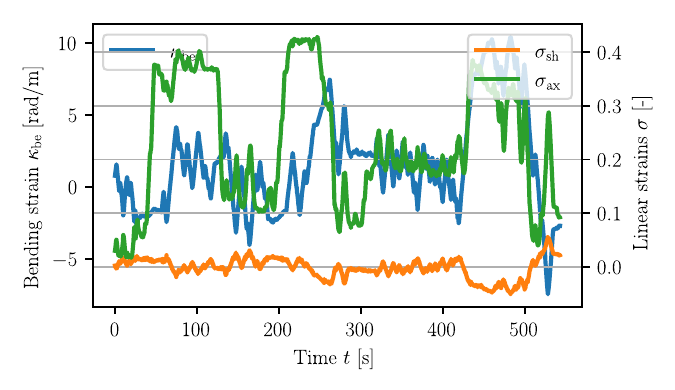}\label{fig:experimental_results:setpoint_regulation:brain:q}}
    \subfigure[Control input $\phi$]{\includegraphics[width=0.47\textwidth, trim={5, 5, 5, 5}]{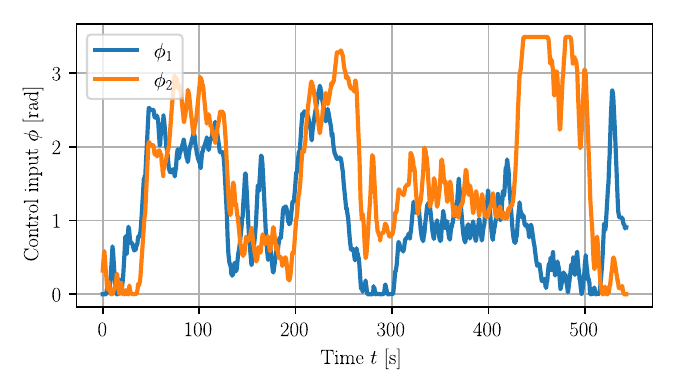}\label{fig:experimental_results:setpoint_regulation:brain:phi}}
    \caption{Experimental results for tracking a reference trajectory of nine step functions with motor imagery. \textbf{Panel (a) \& (b):} The x/y-coordinate of the end-effector position with the solid line denoting the actual position, the dotted line the attractor position, and the dashed line the reference (i.e., the setpoint).
    \textbf{Panel (c):} The evolution of the configuration.
    \textbf{Panel(d):} The saturated planar control inputs. }\label{fig:experimental_results:setpoint_regulation:brain}
\end{figure*}

\subsubsection{Mapping to Lagrangian forces}
Now that we have formulated our control law $\tau$ in configuration space, we need to identify a strategy to specify the motor angles $\phi \in \mathbb{R}^2$ such that $\alpha(q,\phi) \approx \tau$. Note that, in contrast to other continuum soft robots studied in literature~\cite{della2023model}, the actuation term $\alpha(q,\phi)$ is not affine in control. 
In previous work~\cite{stolzle2023experimental}, we side-stepped this challenge by linearizing with respect to the steady-state actuation $\phi^\mathrm{ss}$: $A(q) = \lVert \frac{\partial \alpha}{\partial \phi}\rVert_{\phi=\phi^\mathrm{ss}}$ therefore recovering the usual scenario of an affine actuation function. Unfortunately, this is not possible in the setting of this work as i) we do not have access to such $\phi^\mathrm{ss}$, and ii) linearizing around $\phi$ causes the closed-loop system to become unstable. We, therefore, propose to formulate instead a nonlinear least-squares problem $\phi^\mathrm{d} = \argmin_\phi \frac{1}{2} \lVert \tau - \alpha(q,\phi) \rVert^2$ and solve it in real-time with a Levenberg Marquardt solver implemented in JAX~\cite{jaxopt_implicit_diff}.

We note that this approach is not guaranteed to be valid for the general case of an underactuated soft robot but for this particular structure of $\alpha(q,\phi) \in \mathbb{R}^3$ with $\phi \in \mathbb{R}^2$ it is possible to identify solutions $\phi$ with the Euclidean norm of the residual being smaller than $0.001$.
The source code of the controller is available on GitHub\footnote{\url{https://github.com/tud-phi/hsa-planar-control}}.

\section{Experiments}

\subsection{Experimental setup}
In the following, we detail the \ac{EEG} data processing procedure (see also Fig.~\ref{fig:eeg_pipeline}) and present our experimental setup, which is annotated in Fig.~\ref{fig:experimental_setup}.

\subsubsection{EEG data processing}\label{ssub:eeg_pipeline}
We integrate the 3-channel flexEEG Neuroconcise device with the OpenVibe software
to acquire the EEG data and process it in real-time.
This configuration facilitates data recording and cue presentation. We process the \ac{EEG} signals at a sampling frequency of \SI{125}{Hz} with a pipeline that involves three bi-polar channels around the motor cortex: FC3-CP3, FCZ-CPZ, and FC4-CP4.
After a notch filter of \SI{50}{Hz}, we apply Independent Component Analysis (ICA) to extract three independent components from the recorded \ac{EEG} data, which is represented by the equation $S(t) = W \, X(t)$, where $S(t)$ are the extracted independent components and $W$ represents the unmixing matrix, allowing us to separate eye blink artifacts in \ac{EEG} signals, which is critical for enhancing the accuracy. 
Subsequently, we apply a Butterworth filter bank to isolate specific frequency bands of interest, including 8-15 \si{Hz} and 15-42 \si{Hz}. 
This enhances the ability to analyze \ac{EEG} data by isolating and examining different frequency band components within the \ac{EEG} signals. Once the signals are filtered in sub-bands and epoched with a duration of \SI{2}{s} and time interval of \SI{0.065}{s}, the features are extracted by the log of the power: $L_i(t) = \log\left(P_i(t)\right)$, where the power $P_i(t) = |E_i(t)|^2$ is represented by square of magnitude of the \ac{EEG} signal $E_i(t)$ at time instance $t$.
These features are then provided to a classifier, which we select as \ac{LDA} due to its simplicity~\cite{lotte2014tutorial}.

We implement a second classifier with the same pipeline, where jaw clenching is provided as a muscle artifact that is classified v/s raw EEG data. 
\subsubsection{Robotic system}
We consider a robot consisting of four \ac{HSA} rods, which were 3D printed via digital projection lithography 
from the photopolymer resin Carbon FPU 50. 
Each \ac{HSA} rod is electrically actuated by a Dynamixel M-28 servo up to a maximum twist angle of $\phi_\mathrm{max} = \SI{3.49}{rad}$.
The robot is attached platform-down to a motion capture cage with eight Optitrack Prime X13 cameras tracking at \SI{200}{Hz} the pose of reflective markers attached to the end-effector of the \ac{HSA} robot.
We estimate the current Cartesian-space velocity of the end-effector with a Savitzky-Golay filter. 
Subsequently, we leverage a closed-form expression of the inverse kinematics of a \ac{CS} model~\cite{stolzle2023experimental} to compute the current configuration $q$ of the robot.
We render an image of the current shape of the robot together with the present end-effector position (white dot), the attractor planned by the user (green square), the operational workspace (grey, see also Fig.~\ref{fig:hsa_workspace}) and if applicable, the goal position (red dot). 
We specify the currently active axis of movement $e_\mathrm{a}$ with a double arrow and project the resulting image onto a black screen in the background of the motion capture cage.
The robot is operated with a ROS2 software framework\footnote{\url{https://github.com/tud-phi/ros2-hsa}}. We receive the predicted and classified brain signals via TCP at a frequency of \SI{18}{Hz} and move the attractor subsequently with $\Delta_\mathrm{x} = \SI{0.2}{mm}$. We evaluate the Cartesian impedance controller using the gains $K_\mathrm{p} = \SI{300}{N \per m}$, $K_\mathrm{d} = \SI{1.5}{N s \per \meter}$ at a frequency of \SI{50}{Hz} and finally send the desired motor positions to the servos.

The entire control pipeline from measuring \ac{EEG} signals to sending the actuation commands to the motors exhibits a maximum latency of (i.e., is upper-bounded by)  \SI{130}{ms}.

\subsection{BMI protocol:}\label{sub:bmi_protocol}
In the following, we will describe the protocol for collecting the motor imagery dataset, training the \ac{EEG} classifiers, and mapping classifier outputs into actionable robot commands.

\subsubsection{Training protocol}

The participant is given brief instructions about motor imagery signals.
We follow the Graz-BCI~\cite{roc2021review} paradigm, which assists with training, where the display of the cue instructs the participant to perform imagination of movements: when a left arrow appears, the participant is asked to rest and not focus on motor movement. When the right arrow appears on the screen, the participant is asked to imagine the motor activity from the dominant hand (here, the right hand), such as grasping or squeezing an object. The training protocol for right-hand motor imagery v/s rest demands fifty cues per class. 
We similarly collect data for the second classifier by asking participants to clench their jaw, which can be detected as muscular artifacts (vs. no muscular movement) in the \ac{EEG} signals.
At the end of the trial, we train both classifiers (see Sections~\ref{sub:motor_imagery_bmi} and \ref{ssub:eeg_pipeline} for more information) and repeat the procedure until an accuracy of \SI{75}{\percent} is obtained for right-hand motor imagery v/s rest and accuracy of \SI{85}{\percent} for jaw clench artifact v/s raw EEG.

\begin{figure*}[t]
    \centering
    \subfigure[End-effector x-coordinate]{\includegraphics[width=0.47\textwidth, trim={5, 5, 5, 5}]{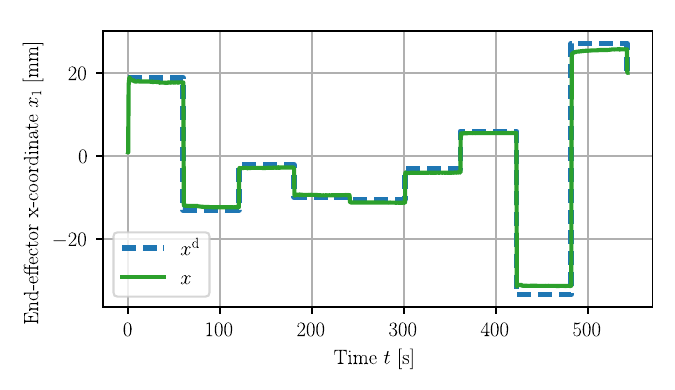}\label{fig:experimental_results:setpoint_regulation:computational:pee_x}}
    \subfigure[End-effector y-coordinate]{\includegraphics[width=0.47\textwidth, trim={5, 5, 5, 5}]{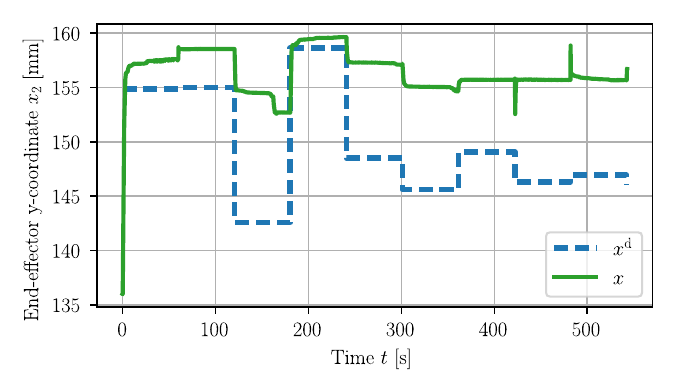}\label{fig:experimental_results:setpoint_regulation:computational:pee_y}}\\
    \vspace{-0.2cm}
    \subfigure[Configuration $q$]{\includegraphics[width=0.47\textwidth, trim={5, 5, 5, 5}]{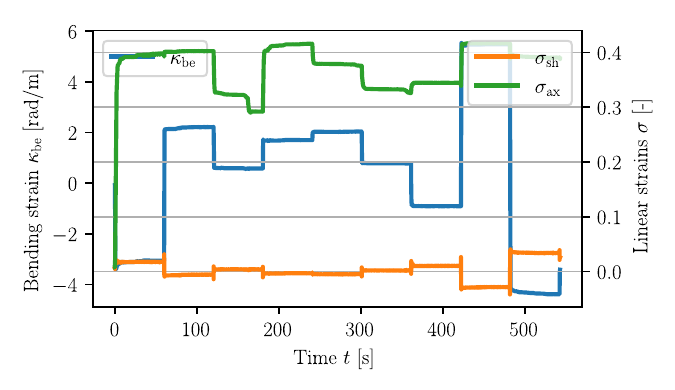}\label{fig:experimental_results:setpoint_regulation:computational:q}}
    \subfigure[Control input $\phi$]{\includegraphics[width=0.47\textwidth, trim={5, 5, 5, 5}]{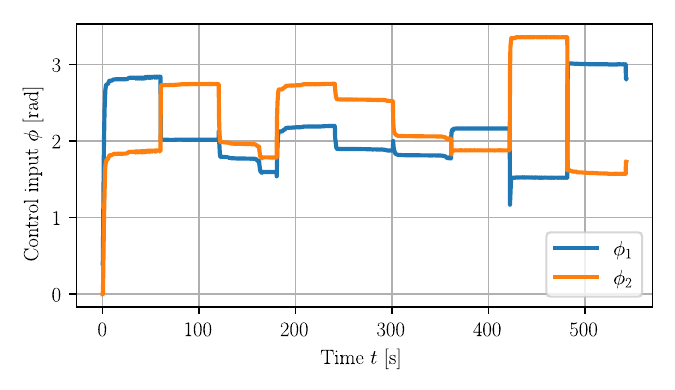}\label{fig:experimental_results:setpoint_regulation:computational:phi}}
    \caption{Experimental results for tracking a reference trajectory of nine step functions directly with the Cartesian impedance controller with access to the privileged information $x^\mathrm{d}$. \textbf{Panel (a) \& (b):} The x/y-coordinate of the end-effector position with the solid line denoting the actual position, the dotted line the attractor position, and the dashed line the reference (i.e., the setpoint).
    \textbf{Panel (c):} The evolution of the configuration.
    \textbf{Panel(d):} The saturated planar control inputs. }\label{fig:experimental_results:setpoint_regulation:computational}
    \vspace{-0.2cm}
\end{figure*}

\subsubsection{Online robot control}
Now, the participant operates the HSA robot in real time, with both classifiers being executed online.
Moreover, we map the outputs of the classifier into actionable commands: we initialize the x-axis as the active axis of movement (i.e., $e_\mathrm{a} = [1,0]$). When the first classifier detects jaw clenching for at least \SI{80}{\percent} of samples over a duration of \SI{2.8}{s}, we switch to the y-axis: $e_\mathrm{a} = [0,1]$ and vice-versa. If we do not detect any jaw clenching artifacts, we evaluate the output of the second classifier in parallel: if there is motor imagery activity identified in the \ac{EEG} classifier, the attractor will be moved in the positive direction (i.e., $s=1$) along $e_\mathrm{a}$. In contrast, if the \ac{EEG} signals are classified as the participant being at rest, $s=-1$ (i.e., movement in the negative direction).


\subsection{Setpoint regulation}\label{sub:experiments:setpoint_regulation}
We randomly generate nine setpoints $x^\mathrm{d}(t) \in \mathbb{R}^2$ within the operational workspace of the robot (see Fig.~\ref{fig:hsa_workspace} and display them as a red circle to the user over a duration of \SI{540}{s}.
The user can freely move the attractor to reach and keep the end-effector at the setpoint.
Furthermore, we execute an experiment in which the computational controller has access to normally privileged information: as we substitute $x^\mathrm{at}$ with $x^\mathrm{d}$ in \eqref{eq:cartesian_impedance_controller}, the Cartesian impedance controller now immediately regulates the robot towards the setpoint.
By excluding the \ac{BMI} from the pipeline, this provides us with a reference of the performance we can expect from the computational controller and, with that, also represents a performance upper bound.

\begin{figure*}[tb]
    \centering
    \subfigure[$t=\SI{0}{s}$]{\includegraphics[width=0.192\textwidth]{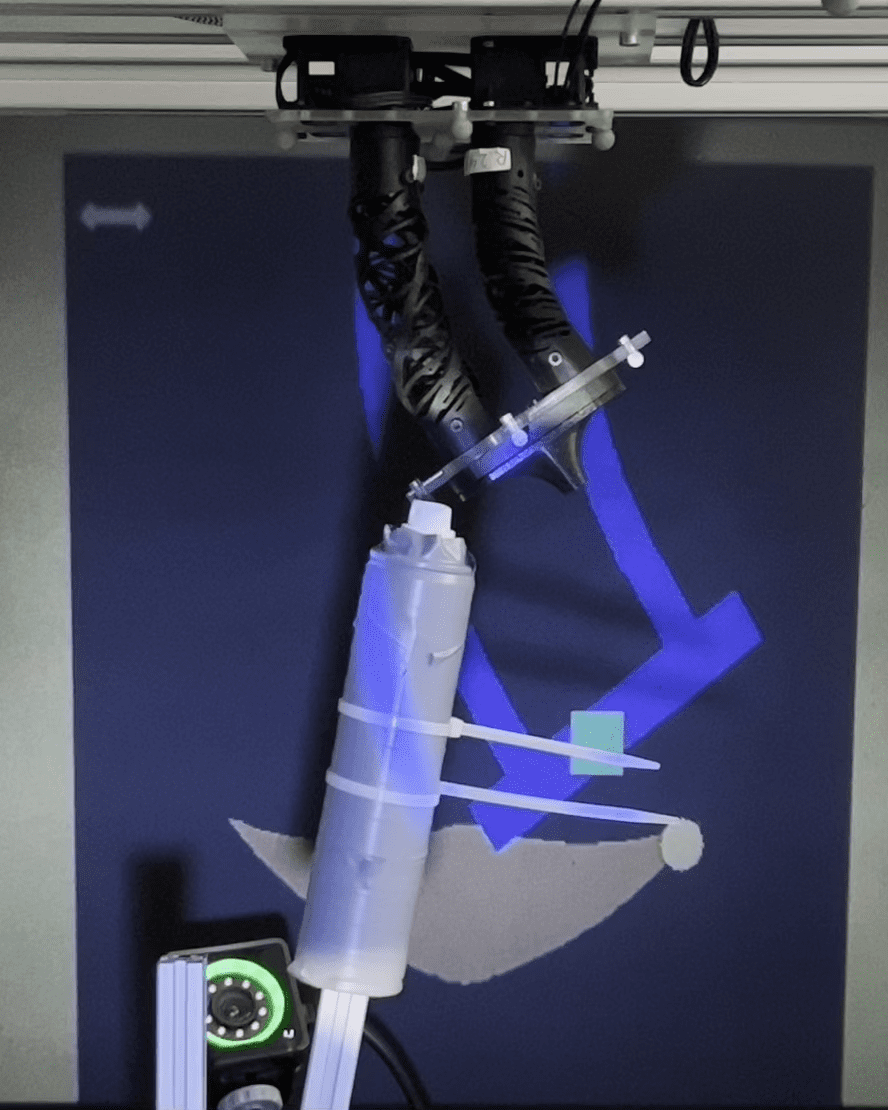}}
    \subfigure[$t=\SI{12}{s}$]{\includegraphics[width=0.192\textwidth]{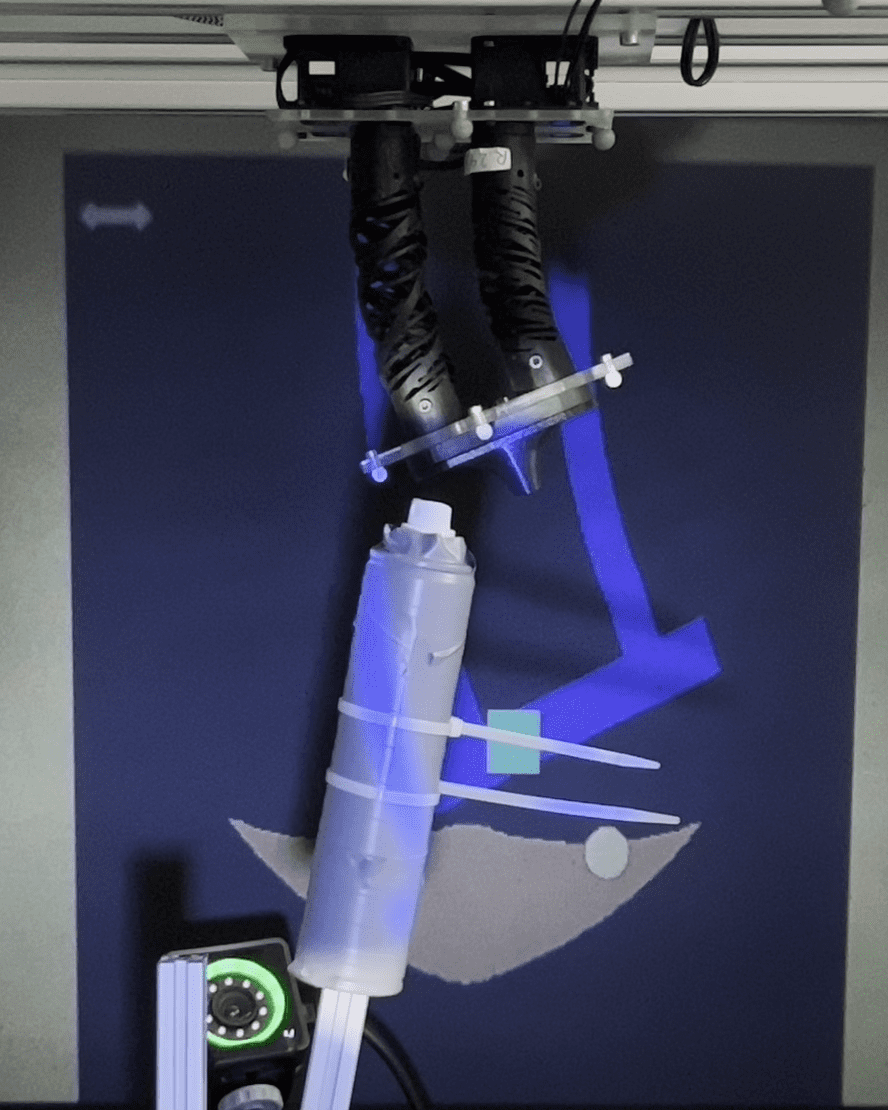}}
    \subfigure[$t=\SI{24}{s}$]{\includegraphics[width=0.192\textwidth]{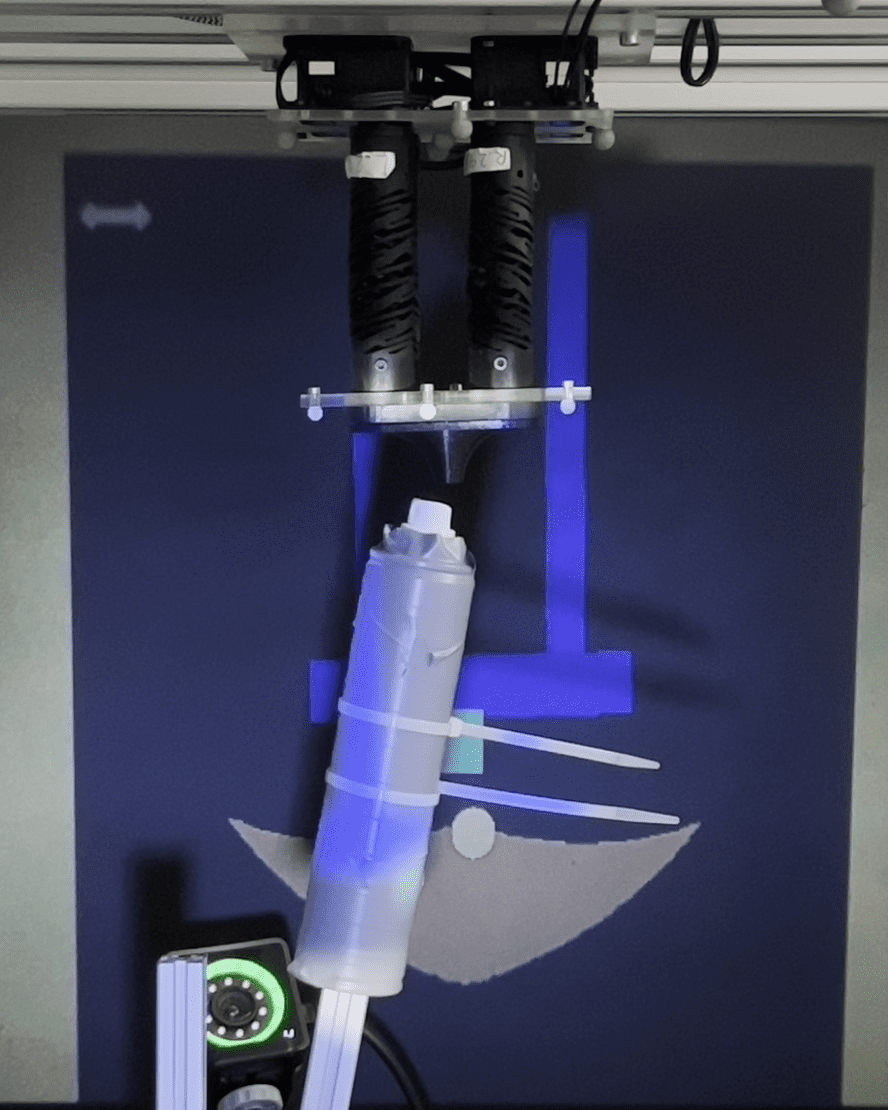}}
    \subfigure[$t=\SI{36}{s}$]{\includegraphics[width=0.192\textwidth]{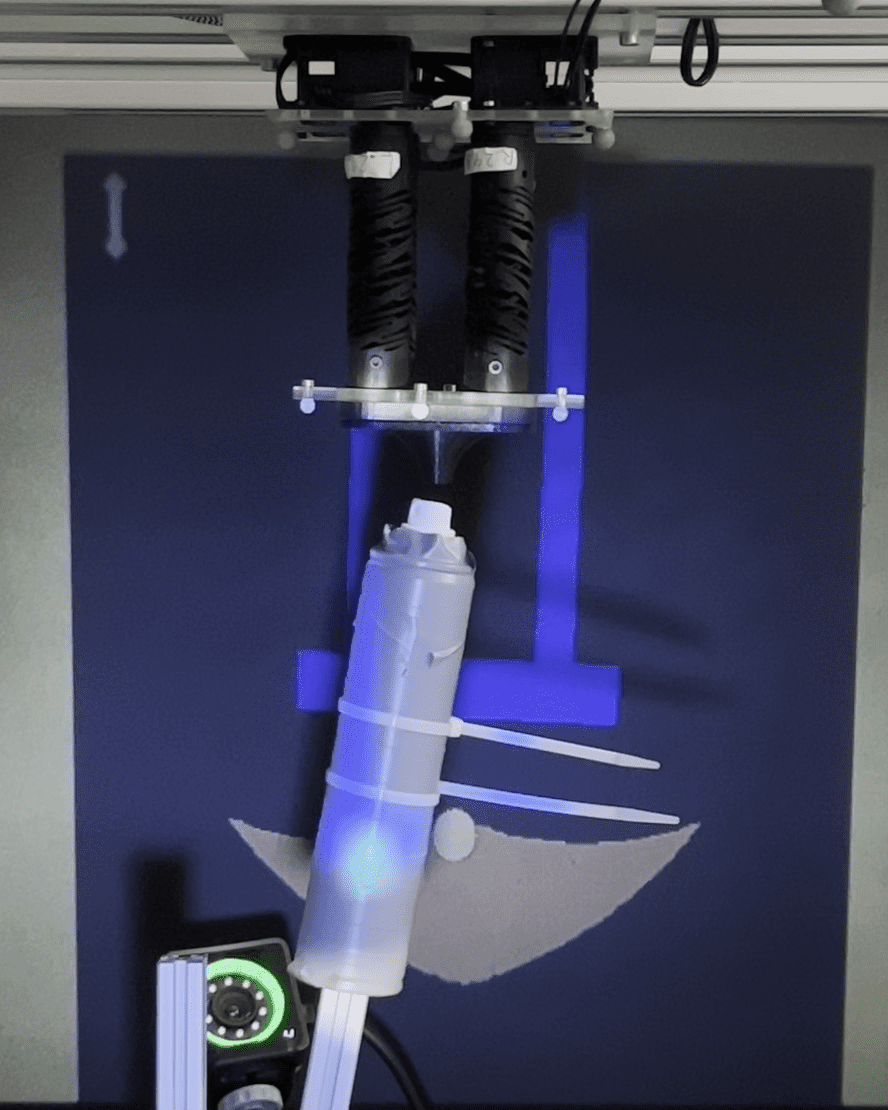}}
    \subfigure[$t=\SI{48}{s}$]{\includegraphics[width=0.192\textwidth]{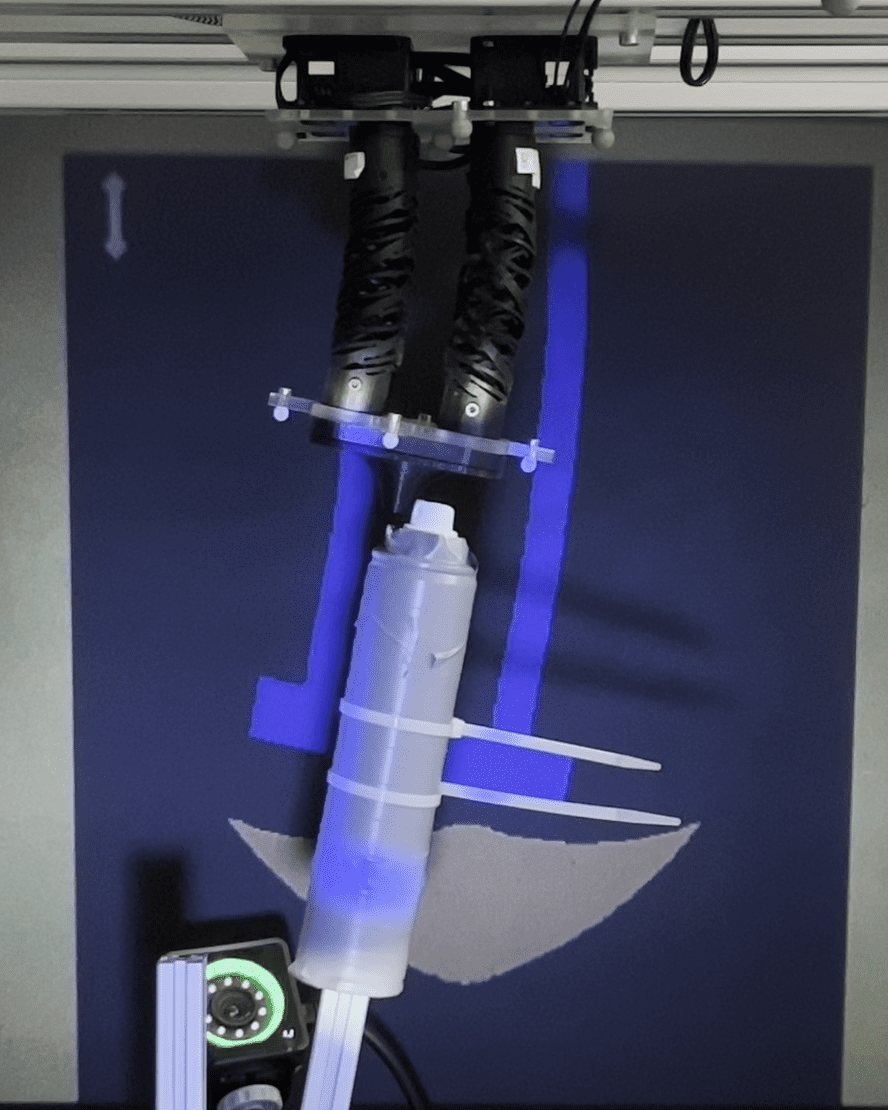}}\\
    \subfigure[$t=\SI{60}{s}$]{\includegraphics[width=0.192\textwidth]{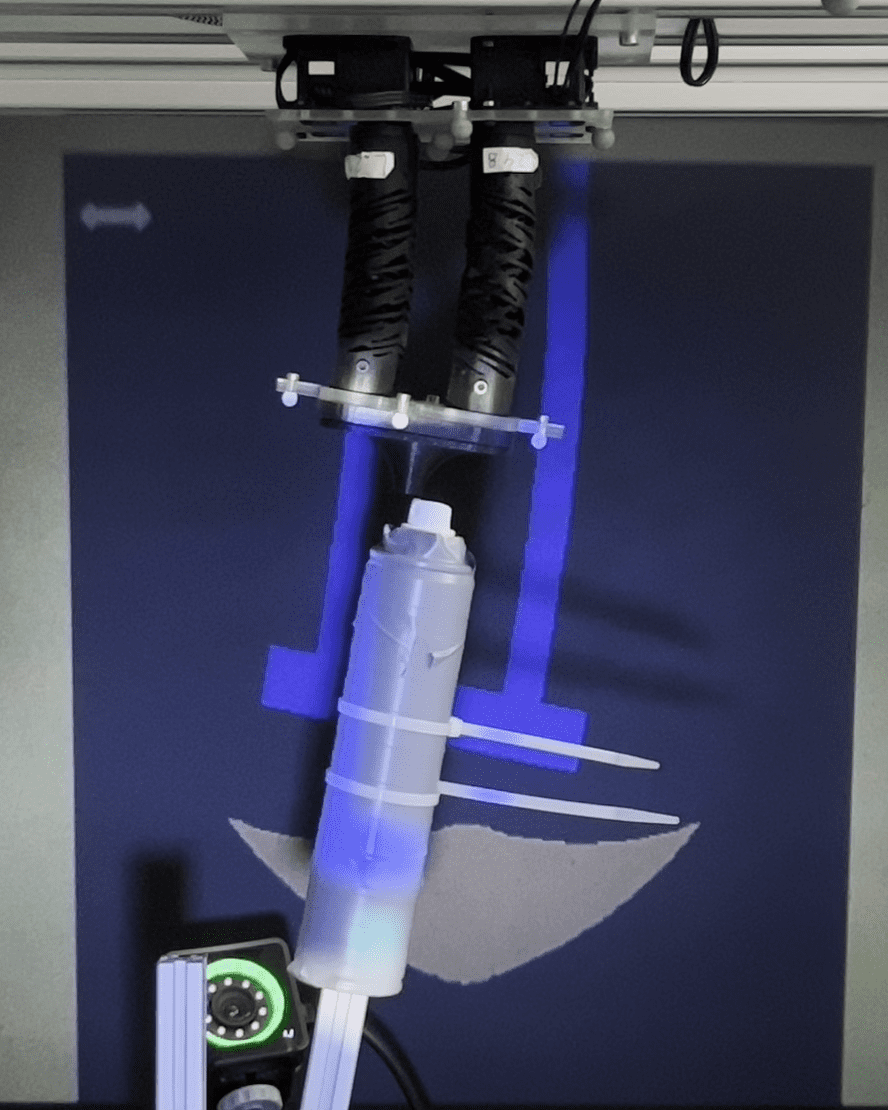}}
    \subfigure[$t=\SI{72}{s}$]{\includegraphics[width=0.192\textwidth]{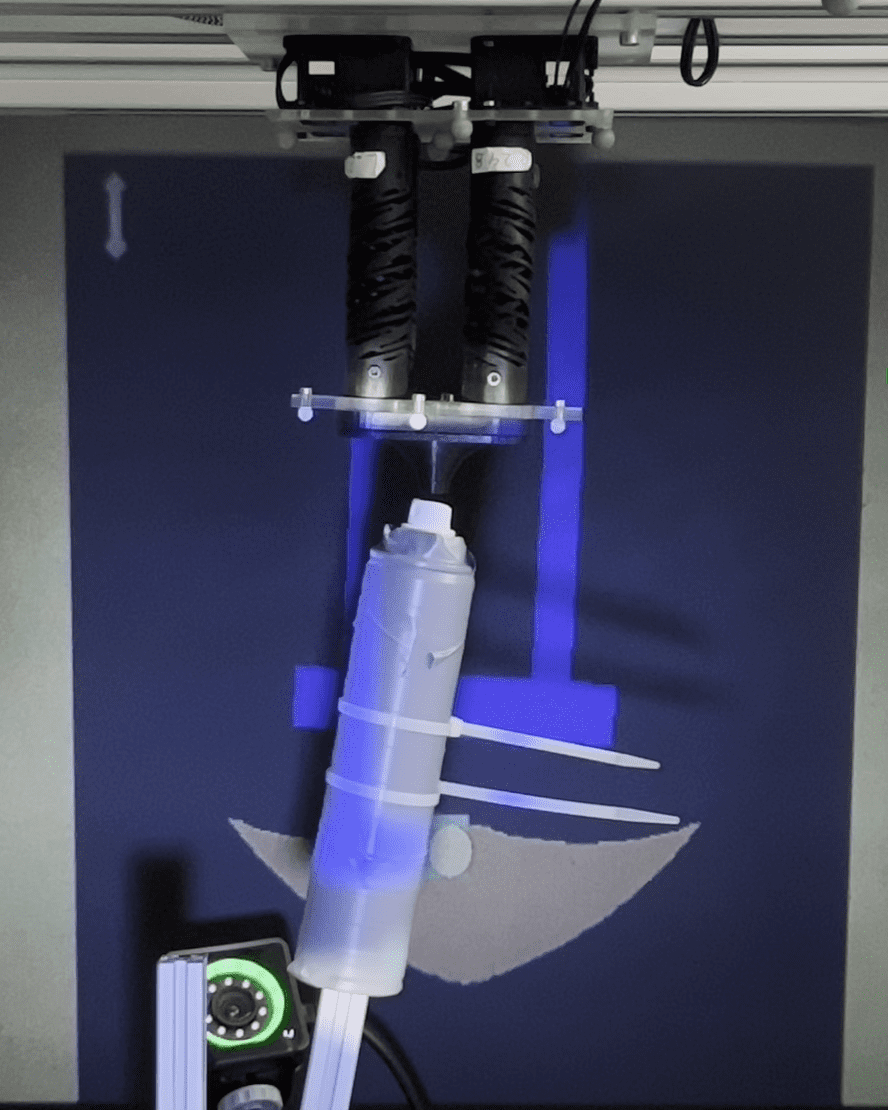}}
    \subfigure[$t=\SI{84}{s}$]{\includegraphics[width=0.192\textwidth]{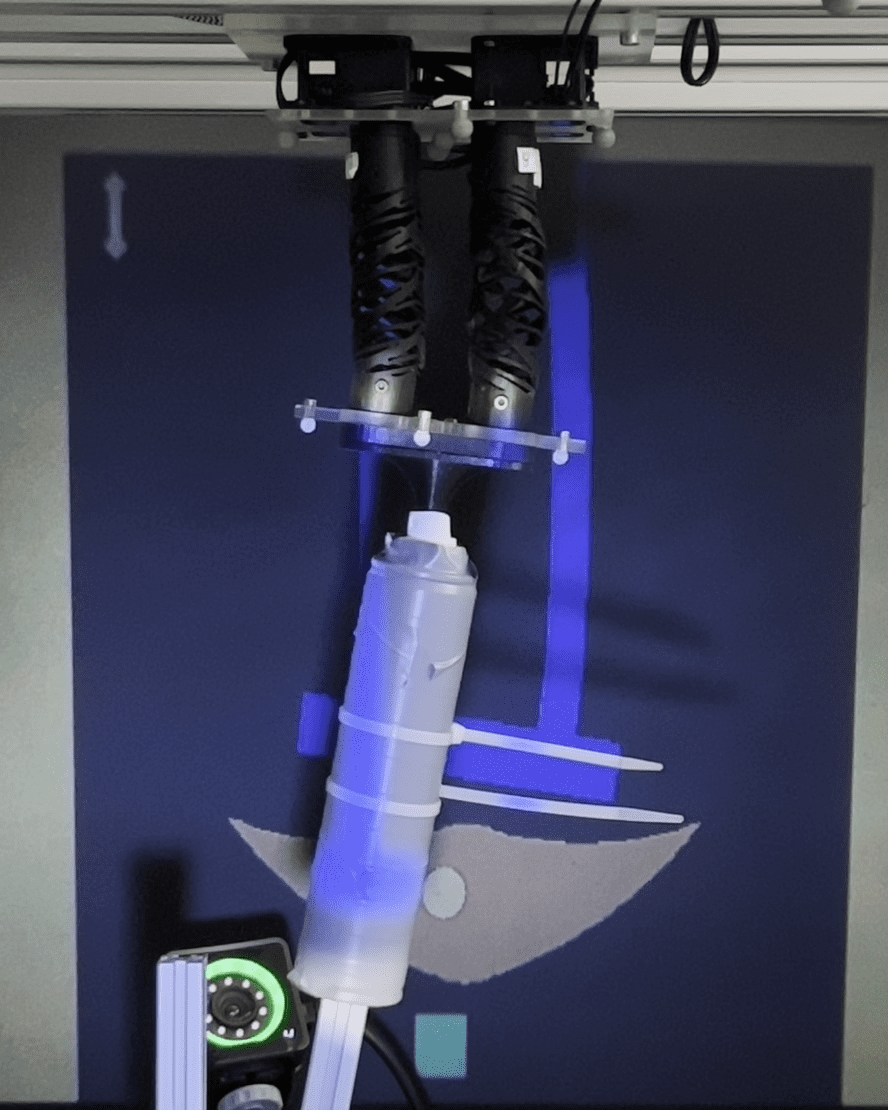}}
    \subfigure[$t=\SI{96}{s}$]{\includegraphics[width=0.192\textwidth]{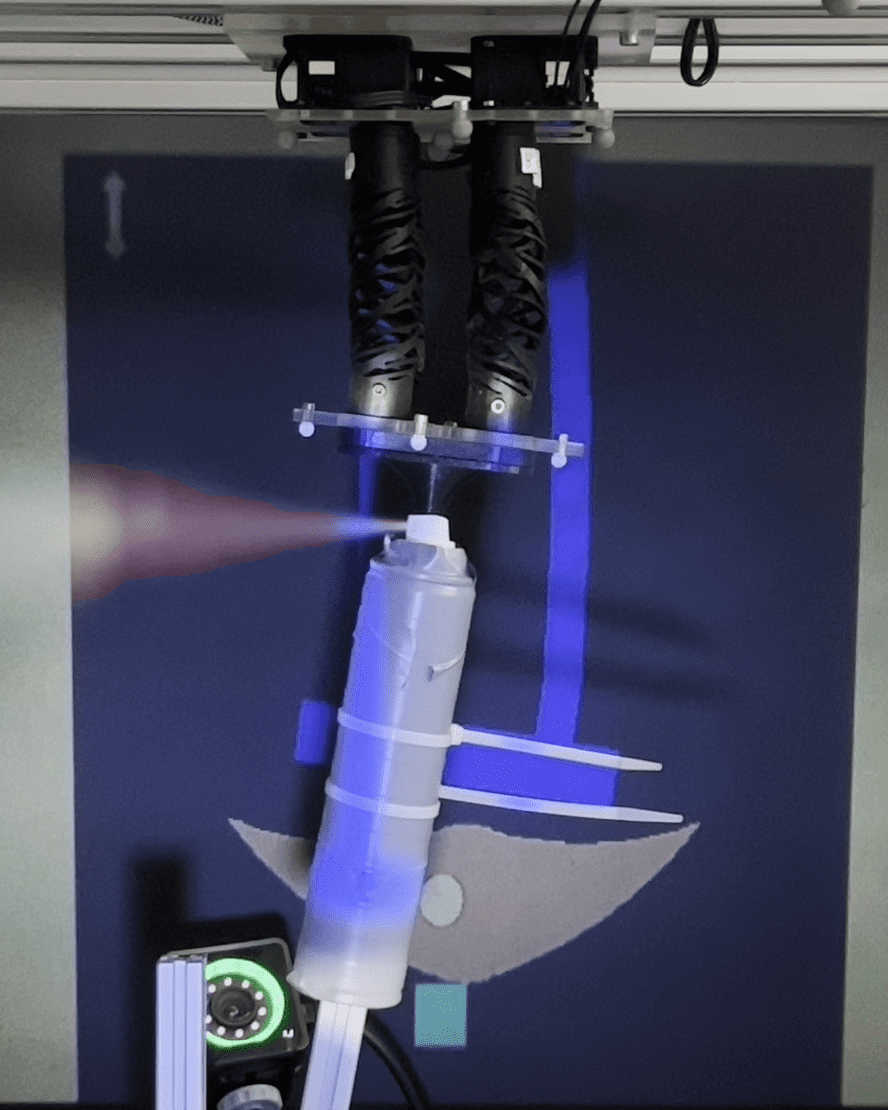}\label{fig:experimental_results:adl_task_sequence_of_stills:spraying}}
    \subfigure[$t=\SI{108}{s}$]{\includegraphics[width=0.192\textwidth]{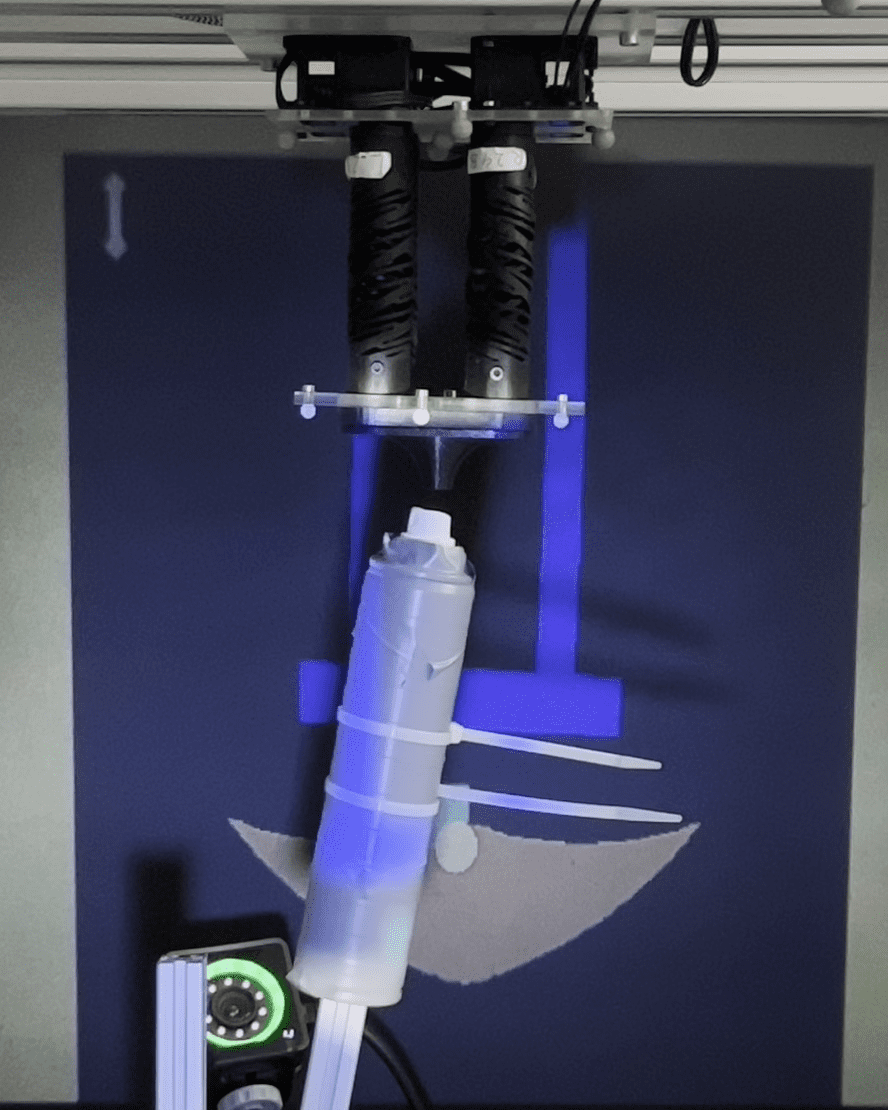}}
    \vspace{-0.2cm}
    \caption{Sequence of stills for completing a basic Activity of Daily Living (ADL) by controlling the robot with EEG-based motor imagery.
    \emph{Note: }Fig.~\ref{fig:experimental_results:adl_task_sequence_of_stills:spraying} is edited for improved contrast.
    }\label{fig:experimental_results:adl_task:sequence_of_stills}
\end{figure*}

\subsection{Interacting with the environment on a real-world task}\label{sub:experiments:adl}
We consider the \ac{ADL} task of releasing hairspray by actuating the button of its container with the \ac{HSA} robot's end-effector. For successful execution, the end-effector must be very stiff in the normal direction of the contact. On the other hand, we might want to benefit from the physical intelligence of the system by being relatively flexible in the tangential direction. Therefore, we first define the perpendicular stiffness $k_\perp = \SI{500}{N\per\meter}$ and the tangential stiffness as $k_\parallel = \SI{50}{N \per \meter}$. We assume that the normal direction of the contact can be described by the polar angle $\theta_\perp$ (with respect to the x-axis). We envision that in the future, the user can adjust such stiffness characteristics online via a \ac{BMI} system similar to~\cite{schiatti2017soft}. In this work, however, we estimate by visual inspection that $\theta_\perp=\SI{1.31}{rad}$.
The Cartesian stiffness matrix in global coordinates is then given by $K_x = R(\theta_\perp) \, \mathrm{diag}(k_\perp, k_\parallel) \, R(\theta_\perp)^\mathrm{T}$
where $R(\theta_\perp) \in SO(2)$ is the rotation matrix between global and contact frames.

\begin{figure*}[tb]
    \centering
    \subfigure[End-effector $x$-coordinate]{\includegraphics[width=0.47\textwidth, trim={5, 5, 5, 5}]{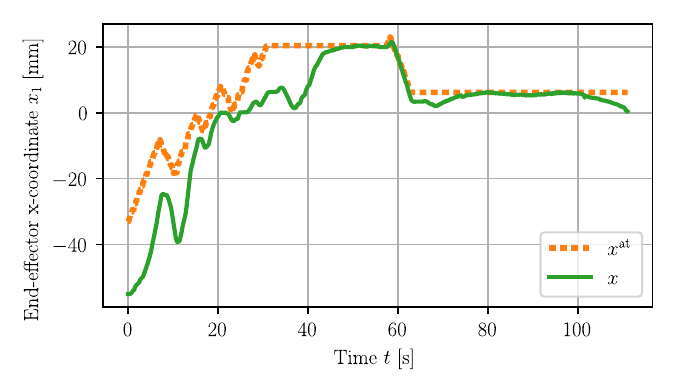}\label{fig:experimental_results:adl_task:brain:pee_x}}
    \subfigure[End-effector $y$-coordinate]{\includegraphics[width=0.47\textwidth, trim={5, 5, 5, 5}]{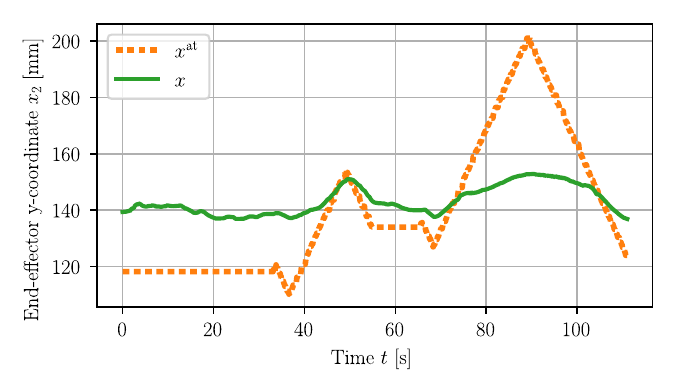}\label{fig:experimental_results:adl_task:brain:pee_y}}
    \\
    \vspace{-0.2cm}
    \subfigure[Configuration $q$]{\includegraphics[width=0.47\textwidth, trim={5, 5, 5, 5}]{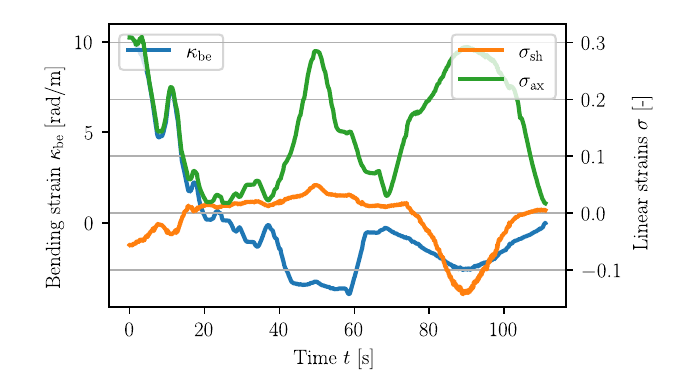}\label{fig:experimental_results:adl_task:brain:q}}
    \subfigure[Control input $\phi$]{\includegraphics[width=0.47\textwidth, trim={5, 5, 5, 5}]{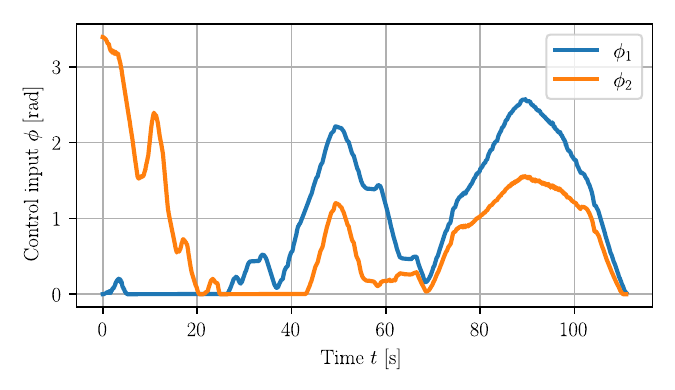}\label{fig:experimental_results:adl_task:brain:phi}}
    \caption{Experimental results for completing a basic Activity of Daily Living (ADL) by controlling the robot with EEG-based motor imagery. \textbf{Panel (a) \& (b):} The x/y-coordinate of the end-effector with the solid line and dotted lines denoting the actual and attractor position, respectively.
    \textbf{Panel (c):} The evolution of the configuration.
    \textbf{Panel(d):} The saturated planar control inputs. }\label{fig:experimental_results:adl_task:brain}
\end{figure*}

\subsection{Evaluation metrics}
In the following, we will introduce and define a few metrics that help us assess the performance of the approach.

\subsubsection{Event-Related Synchronization/De-Synchronization:}
We apply \ac{ERD} / \ac{ERS}~\cite{pfurtscheller1999event} to demonstrate the difference between \ac{EEG} signals when the participant imagines right-hand movement vs. rest, i.e., no activity. \ac{ERD}/\ac{ERS} corresponds to a shift in power during imagination with respect to a baseline. It is defined by
\begin{equation}
    \mathrm{ERD/ERS}(t, f) = \frac{P(t, f) - P_{\mathrm{base}}(f)}{P_{\mathrm{base}}(f)},
\end{equation}
where $\mathrm{ERD/ERS}(t, f)$ represents the \ac{ERD} or \ac{ERS} at a specific time $t$ and frequency $f$, $P(t, f)$ stands for the power of brain activity during imagination, and $P_\mathrm{base}(f)$ denotes the baseline power. 

\subsection{Step response metrics}
For the task of setpoint regulation, we analyze primarily two aspects: (a) is the participant able to reach the proximity of the setpoint within the (generously) allotted time of \SI{60}{s}? We define the proximity of the setpoints as $\lVert x^\mathrm{d} - x(t)\rVert_2 \leq \SI{2}{mm}$. And (b) what is the response time for reaching for the first time the proximity of the setpoint?

\section{Results and Discussion}\label{sec:results_and_discussion}

First, we analyze the \ac{ERD}/\ac{ERS} behavior with respect to rest vs. motor imaginations in Fig.~\ref{fig:ERDS}. It is evident that the baseline of rest remained the same in both scenarios when the participant did not perform motor imagery, but as soon as the cue is presented at \SI{0.0}{s}, a shift in power for the right-hand motor imagery with comparison to rest state is noticeable. 

We present the results for setpoint regulation employing motor imagery in Fig.~\ref{fig:experimental_results:setpoint_regulation:brain}. 
We observe that the participant can reach the proximity of the setpoint within the allotted time of \SI{60}{s} six out of nine times (i.e., \SI{66.6}{\percent}).
For the successful steps, the average response time is \SI{21.5}{s}.
However, as our protocol does not contain a command to let the attractor rest, it is challenging to keep the end-effector at the setpoint, and we observe oscillations, particularly with respect to the x-coordinate.
In our third experiment, we ask the Cartesian impedance controller to track the setpoints directly.
The fast response time, a well-known characteristic of model-based control approaches, is evident. However, the errors in the model (for example, caused by hysteresis or unmodelled nonlinearities)~\cite{stolzle2023experimental}, together with the lack of integral action, lead to steady-state errors. 

Finally, we consider the \ac{ADL} task of releasing hair spray using the end-effector of the \ac{HSA} robot. We present a sequence of stills in Fig.~\ref{fig:experimental_results:adl_task:sequence_of_stills} and plots of the entire sequence in Fig.~\ref{fig:experimental_results:adl_task:brain}.
Already during the first attempt, the participant can steer the end-effector toward the button, apply force, and release the fluid within \SI{86}{s}.
The impedance of the controller is clearly visible in Fig.~\ref{fig:experimental_results:adl_task:brain:pee_y} when the manipulator is in contact with the object at time \SI{74}{s} to \SI{104}{s}.
Also, we noticed that the end-effector does not need to be perfectly aligned with the center of the button and can still complete the task successfully due to the compliance of the closed-loop system in the tangential direction.

We noticed that the variability of setting up the \ac{EEG} device on each study participant and the \ac{EEG} sensor noise caused by external factors (e.g., floor vibrations) still pose a considerable challenge for deploying motor imagery-based tools in practice. Furthermore, subject-specific factors such as the ability to focus on imagining motor actions, mental tiredness, etc., significantly affected the performance (e.g., classification accuracy, setpoint tracking error).

\section{Conclusion}
In this paper, we propose to combine motor imagery-based \ac{BMI} systems with continuum soft robots. This symbiosis promises the safe and compliant operation of robots that can assist people with limb impairments in their daily lives.
As demonstrated in the \ac{ADL} experiment, the physical intelligence of the soft robot can compensate for errors and deviations in the output of the \ac{BMI} classifier.
Furthermore, we introduced a Cartesian impedance controller for planar \ac{HSA} robots that can deal with the peculiar characteristics of these robots (e.g., underactuation, non-affinity in control, etc.), and allows for model-based control without interfering with the structural compliance of the system.

\section*{Acknowledgment}
The authors would like to thank Dr. Fabien Lotte for his suggestions concerning the protocol, Dr. Tomas Ward and the Neuroconcise team for their support with the FlexEEG device, and J.K. Balasubramanian for his assistance with the EEG setup.

\bibliographystyle{IEEEtran}
\bibliography{main}

\end{document}